%% file: example_paper.tex
\documentclass{article}

\usepackage{microtype}
\usepackage{graphicx}
\usepackage{subfigure}
\usepackage{booktabs} 

\usepackage{hyperref}

\usepackage[accepted]{icml2024}

\usepackage{amsmath}
\usepackage{amssymb}
\usepackage{mathtools}
\usepackage{amsthm}

\usepackage[capitalize,noabbrev]{cleveref}

\theoremstyle{plain}

\theoremstyle{definition}

\theoremstyle{remark}

\usepackage[textsize=tiny]{todonotes}

\icmltitlerunning{Latent Diffusion Model for Generating Ensembles of Climate Simulations}

\begin{document}

\twocolumn[
\icmltitle{Latent Diffusion Model for Generating Ensembles of Climate Simulations}




\begin{icmlauthorlist}
\icmlauthor{Johannes Meuer}{yyy}
\icmlauthor{Maximilian Witte}{yyy}
\icmlauthor{Tobias Sebastian Finn}{zzz}
\icmlauthor{Claudia Timmreck}{xxx}
\icmlauthor{Thomas Ludwig}{yyy}
\icmlauthor{Christopher Kadow}{yyy}
\end{icmlauthorlist}

\icmlaffiliation{yyy}{Data Analysis, German Climate Computing Center (DKRZ), Hamburg, Germany}
\icmlaffiliation{zzz}{CEREA Écoles des Ponts and EDF R\&D, Paris, France}
\icmlaffiliation{xxx}{Max-Planck Institute of Meteorology, Hamburg, Germany}

\icmlcorrespondingauthor{Johannes Meuer}{meuer@dkrz.de}

\icmlkeywords{Machine Learning, ICML}

\vskip 0.3in
]



\printAffiliationsAndNotice{}  

\begin{abstract}
Obtaining accurate estimates of uncertainty in climate scenarios often requires generating large ensembles of high-resolution climate simulations, a computationally expensive and memory intensive process. To address this challenge, we train a novel generative deep learning approach on extensive sets of climate simulations. The model consists of two components: a variational autoencoder for dimensionality reduction and a denoising diffusion probabilistic model that generates multiple ensemble members. We validate our model on the Max Planck Institute Grand Ensemble and show that it achieves good agreement with the original ensemble in terms of variability. By leveraging the latent space representation, our model can rapidly generate large ensembles on-the-fly with minimal memory requirements, which can significantly improve the efficiency of uncertainty quantification in climate simulations.
\end{abstract}

\section{Introduction}
\label{introduction}

Climate simulations are essential tools for understanding Earth system processes and supporting diverse applications. However, these simulations exhibit internal variability due to chaotic variability and unknown forcings. Ensemble-based approaches, such as the Max Planck Institute (MPI) Grand Ensemble \cite{maher2019max}, address these uncertainties by providing a collection of simulations with varied initial conditions and model parameters. Nevertheless, these ensembles are computationally expensive and often limited in scope.

Machine learning has emerged as a promising complementary tool, capable of uncovering patterns and correlations. Reichstein et al. \citeyearpar{reichstein2019deep} discuss the growing role of deep learning in improving climate science, highlighting its potential to identify non-linear relationships between climate variables. Their work illustrates how deep learning can reveal previously hidden patterns that improve climate simulations. Ensemble-based learning, as explored by \cite{lorenz2018prospects}, further improves predictive accuracy by weighting climate models based on their historical performance. This approach allows better integration of the strengths of different models, leading to more robust predictions.

Recent work using generative adversarial networks (GANs) shows promise in climate modelling by generating realistic weather simulations that match high-resolution numerical models. Brochet et al. \citeyearpar{brochet2023multivariate} demonstrate how GANs can provide multivariate emulation of numerical weather predictions. However, while GANs are powerful in generating plausible simulations, ofthen they suffer from mode collapse, where the generator produces limited types of output and fails to cover the diversity of the training data. This deficiency is particularly problematic in climate modelling, where robust sampling from the distribution of climate simulations is crucial for uncertainty quantification.

\begin{figure*}[ht]
\vskip 0.2in
\begin{center}
\centerline{\includegraphics[width=\textwidth]{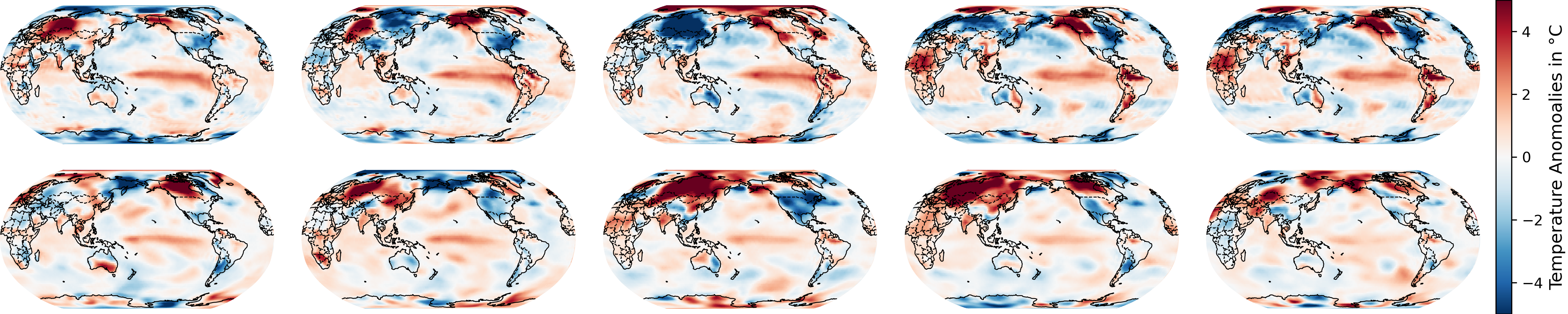}}
\caption{Monthly anomalies of the strongest El Niño events. Top row shows an original simulation of the MPI-GE, bottom row a selected generated simulation using our latent diffusion model.}
\label{enso-maps}
\end{center}
\vskip -0.2in
\end{figure*}

Denoising Diffusion Models \cite{ho2020denoising} solve this problem by sampling from a Gaussian noise distribution and minimising the KL divergence between the distribution of the predictions and the training data. This allows stable training and effective uncertainty quantification. For example, Rasul et al. \citeyearpar{rasul2021autoregressive} use an autoregressive diffusion model to perform multivariate probabilistic forecasting, significantly improving the ability to simultaneously predict different climate-related variables. Another notable application is the use of generative diffusion models to capture the inherent uncertainties in weather forecasting with GenCast by Price et al. \citeyearpar{price2023gencast}. By creating an ensemble of forecasts, GenCast provides a range of probable weather scenarios, which is crucial for medium-range forecasting. Their approach helps in better capturing the variability and uncertainties associated with weather patterns.

We address the challenges of ensemble climate modelling with a machine learning technique based on generative diffusion models. We generate temporally coherent simulations conditioned on a single climate simulation. With this objective, we can efficiently sample an implicit representation of the distribution that specifies the uncertainty conditioned on one climate model simulation. A drawback of diffusion models is their computational time, as the denoising process has to be run iteratively, making it difficult to prarallelize. Furthermore, when used in an auto-regressive prediction model, the computational time scales linearly with the predicted time domain. We present a diffusion model that addresses the efficiency drawbacks of the original diffusion approach by sampling from a latent space. We also introduce two different techniques for generating long sequences: an autoregressive prediction technique that generates long sequences iteratively and a transformer-based technique that can generate long sequences in a single step. Our model successfully reconstructs realistic climate patterns (see figure \ref{enso-maps}) and its simulations provide a similar range of possible outcomes compared to the numerical simulations. Our model exploits the strengths of deep learning, in particular diffusion models, to generate diverse simulations that complement existing ensemble approaches to provide improved uncertainty quantification in climate modelling.

\section{Methodology}

\begin{figure}[h!]
\def\svgwidth{\linewidth}
\centering
\footnotesize{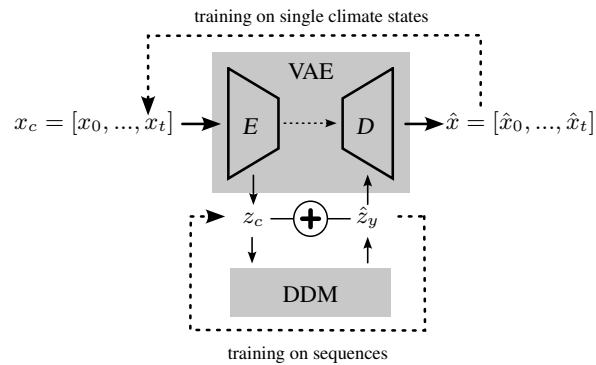}
\caption{Our latent diffusion approach is split into two models, a variational autoencoder (VAE) pre-trained on independent climate states and a denoising diffusion model (DDM) trained on sequences of latent representations. During inference, the DDM generates new simulations in latent space, which are remapped to the original resolution by the decoder (D).}
\label{fig:approach}
\end{figure}

\subsection{Latent Diffusion Model}
Our model (see \ref{fig:approach}) uses a diffusion process in latent space \cite{rombach2022high} generated by a pre-trained variational autoencoder (VAE) \cite{kingma2013auto}. This approach significantly reduces spatial complexity while preserving essential features of the climate simulations. The pre-trained VAE, described in detail in the appendix \ref{appendix1}, compresses each simulation \(x\) into a lower-dimensional latent space \(z\) using the encoder \(E\):
$$
   z = E(x)
$$
The VAE is unaware of the time dimension and treats each timestep independently. The diffusion model, detailed in Appendix \ref{appendix2}, is trained on the latent representations of the climate simulations, conditioned on a single simulation $x_{c}$ being mapped into latent space, giving a general focus on long-term trends in climate evolution: \(z_{c} = E(x_{c})\). The prediction task is defined as the difference between a target latent \(z\) and the conditioned latent simulation \(z_{c}\), where \(z_{y}\) represents the residual that the diffusion model has to learn:
$$
z_{y} = z - z_{c}
$$
During training, the diffusion model is optimised to predict this residual in latent space. The diffusion model is a U-Net \cite{ronneberger2015u} using BigGAN \cite{brock2018large} residual blocks followed by down-sampling convolutions in the encoder and up-sampling convolutions in the decoder. After training, during inference, we use the diffusion model (DDM) to generate a large number of residuals \(\hat{z}_{y}\) in the latent space:
$$
\hat{z}_{y} = \text{DDM}(z_{c})
$$
To speed up the inference time of the diffusion model, we apply a denoising diffusion implicit sampler \cite{song2020denoising}, which provides a more efficient generation process by using deterministic steps. The final simulations are reconstructed by adding the generated residuals back to the conditioned latent simulation $z_{c}$ and applying the VAE decoder $D$:
$$
\hat{x} = D(z_{c} + \hat{z}_{y})
$$

\subsection{Sequence Generation}

We explore two approaches to generating long sequences in the latent space:

\subsubsection{Autoregressive Prediction}
This approach iteratively generates long sequences by predicting the next latent state based on a window of previous states and the conditioned simulation $z_{c}$. Given an input sequence \(z_{t-n+1:t} = [z_{t-n+1}, z_{t-n+2}, \ldots, z_{t}] \), the model predicts:

$$
   \hat{z}_{t+1} = DDM(z_{t-n+1:t}, z_{c})
$$

\subsubsection{Transformer-Based Attention Mechanism}
Inspired by natural language processing \cite{vaswani2017attention}, this approach uses a transformer to process the entire time domain at once. This allows parallel processing and accelerates sequence generation. Each transformer block sequentially applies spatial attention, focusing on spatial patterns, and temporal attention, focusing on temporal correlations, following a residual block. To manage memory costs, we implement a cascaded transformer mechanism. The higher levels of the diffusion network focus on small time scales, capturing detailed short-term patterns. The lower levels deal with overall time scales, ensuring a comprehensive understanding of long-term trends. A detailed description of the model can be found in Appendix \ref{appendix2}. The transformer processes the entire sequence in a single step and is not additionally conditioned on an initial state:
$$
   \hat{z} = DDM(z_{c})
$$


\section{Results}
Our model is trained and evaluated on the 200 ensemble members from the historical simulations of the MPI Grand Ensemble \cite{maher2019max}. This ensemble covers the period from 1850 to 2005 with a monthly temporal frequency and a spatial resolution of 1.8° (192x96 grid). The focus of this analysis is on surface temperatures.

\begin{figure}[ht]
\vskip 0.2in
\begin{center}
\centerline{\includegraphics[width=\columnwidth]{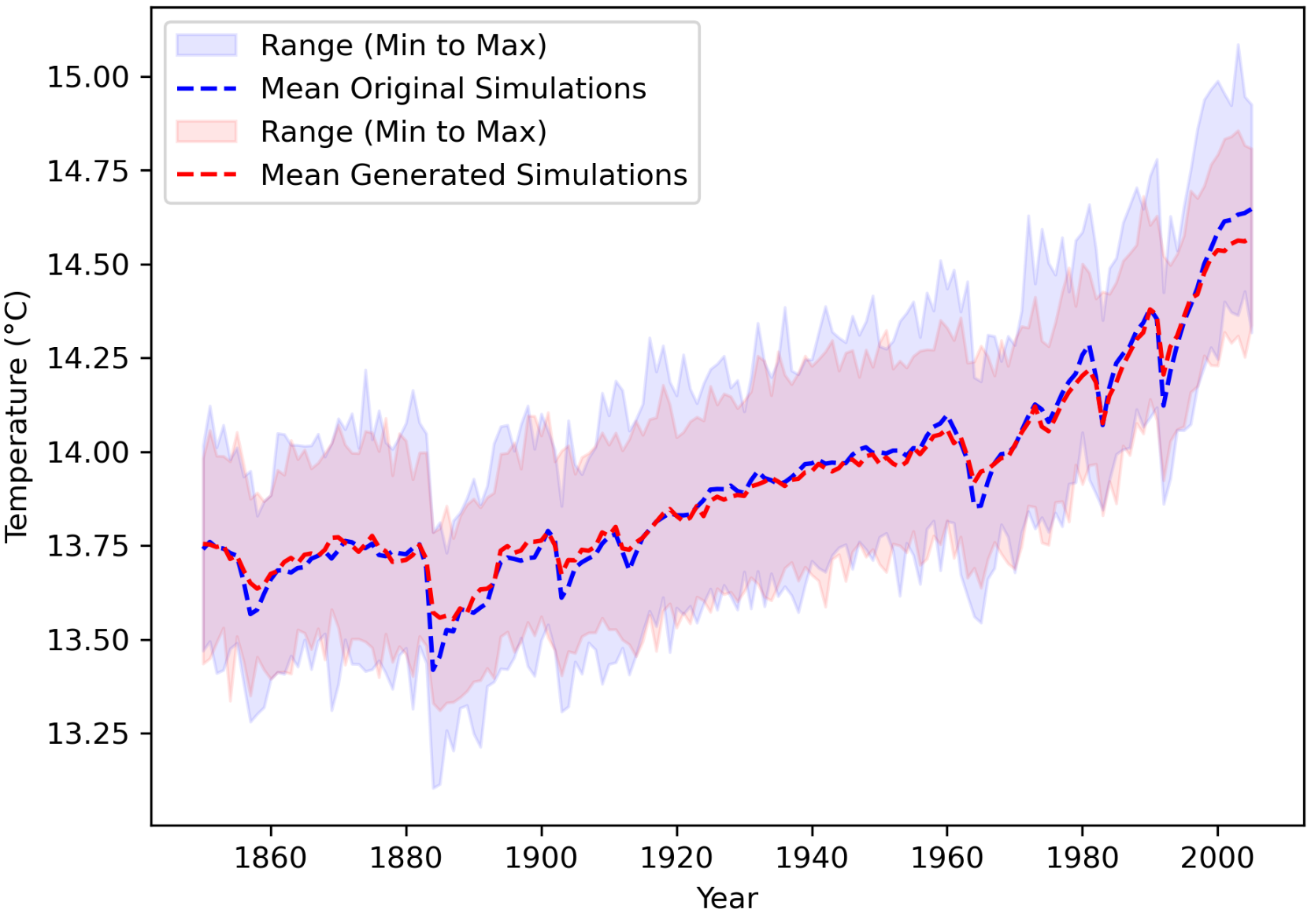}}
\caption{Ensemble spread and ensemble mean of annual spatially averaged 98 original members from the MPI-GE (blue) compared to the generated members (red) from 1850 to 2005.}
\label{results-mean-spread}
\end{center}
\vskip -0.2in
\end{figure}

During training, we used one member as input for conditional simulation. For model evaluation, we used another member for conditioning during inference. We trained on the remaining 198 members over the entire time range. After training, we generated 100 new artificial ensemble members. These generated members are then analysed to compare their annual mean surface temperatures over the entire historical period with the first 100 members of the original MPI Grand Ensemble. The comparison focuses on two key statistical measures: the ensemble mean and the spread in the temperatures. Figure \ref{results-mean-spread} shows the results of our transformer-based model. While the ensemble mean contains signatures of the forced response to climate change, the ensemble standard deviation represents the internal variability of the climate system.

The ensemble mean and variability of the generated members closely mirror those of the original ensemble members with respect to annual spatially averaged temperatures. This highlights the ability of the model to reproduce central tendencies such as the global warming trend and global cooling following major volcanic eruptions (1883, 1963, 1982 and 1991). This validates the ability of our machine learning approach to reproduce complex climate dynamics. Appendix \ref{appendix3} also provides an analysis of uncertainty quantification for an unseen time range, where the autoregressive model was trained on data from 1850 to 1975 only and iteratively generated simulations for 1975 to 2000.

\begin{figure}[ht]
\vskip 0.2in
\begin{center}
\centerline{\includegraphics[width=\columnwidth]{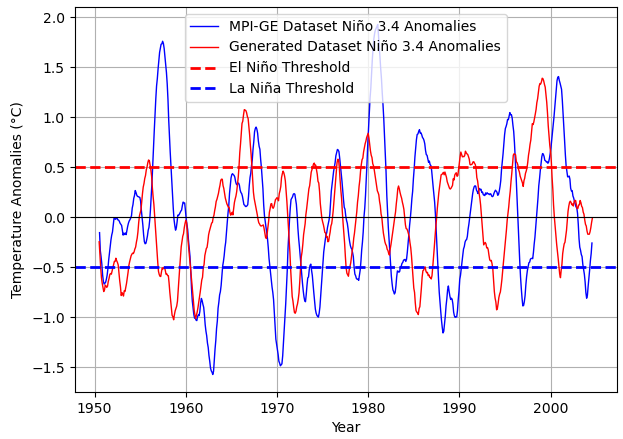}}
\caption{ENSO timeline \cite{trenberth1997definition} of an original simulation from the MPI-GE (blue) in comparison to a generated member (red) ranging from 1950 to 2005. The red dashed line marks the threshold of an El Niño event, the blue dashed line the threshold of a La Niña event.}
\label{enso-timeline}
\end{center}
\vskip -0.2in
\end{figure}

In a further analysis, we looked specifically at the El Niño-Southern Oscillation (ENSO) \cite{trenberth1997definition} timelines of a selected member to assess the model's ability to capture more localised and medium-term climate phenomena. Figure \ref{enso-maps} shows the anomaly maps of the strongest El Niño event from an original simulation and a selected member, which was generated by our autoregressive model. Although the El Niño appears less pronounced in our generated simulation, the model is able to generate temporally and spatially coherent El Niño patterns. This can be also seen in Figure \ref{enso-timeline}, which shows the ENSO timeline from 1950 to 2005. The generated data show a realistic pattern of recurring ENSO events, similar to those observed in real climate data. This similarity confirms that our model not only maintains general climate trends over time, but also effectively reproduces specific, influential climate phenomena such as ENSO.

We found that the autoregressive model was better at preserving the evolution over time, while the transformer model was better at preserving spatial patterns and long-term trends. A comparison of absolute temperature maps can be seen in figure \ref{temperature-maps} in the appendix.

Our results demonstrate that our machine learning framework, integrating a variational autoencoder and a diffusion model, can effectively generate plausible climate scenarios that are statistically consistent with simulated historical data. This capability marks a significant advance in the field of climate modelling, particularly for sampling that accounts for the internal variability and projection of long-term climate phenomena.

\section{Conclusion and Outlook}
We present a latent diffusion model for numerical climate model emulation, highlighting its ability to reproduce both global and local climate phenomena. Our results show that the model effectively captures central tendencies such as the global warming trend and significant cooling events following major volcanic eruptions. Furthermore, the model successfully generates coherent temporal and spatial patterns, such as the El Niño-Southern Oscillation, even when trained on limited historical data.

We carried out a detailed analysis of two different diffusion models: An autoregressive prediction approach and a transformer-based approach. When evaluated on the MPI Grand Ensemble, the generated ensemble members of the transformer-based approach show remarkable agreement with the original ensemble in terms of mean and variability, confirming the robustness and reliability of the model. The autoregressive approach showed the ability to generate realistic climate patterns over extended periods, including unseen time spans, and good performance in temporally coherent simulations. Future work will investigate the combination of the two techniques to leverage the strengths of both.

The promising results of our approach open up several avenues for future research and development. Future work can extend the training data to include more recent years, higher spatial resolutions and multiple climate models. This would allow the model to capture more detailed and recent climate phenomena, improving its applicability to contemporary climate studies. In addition, the inclusion of more climate variables such as precipitation, sea level pressure and ocean currents could provide a more comprehensive understanding of climate dynamics and improve the predictive capabilities of the model. The techniques and insights from this work could be applied to other fields requiring time-series prediction and uncertainty quantification, such as economics, epidemiology and energy systems. A deeper exploration of uncertainty quantification would provide more insight into the confidence and reliability of predictions, which is crucial for policy making and scientific research.

Our latent diffusion model not only quantifies uncertainty, but also provides real scenarios that support the uncertainty quantification. Unlike traditional methods, which often provide abstract uncertainty metrics, our approach generates diverse and plausible climate simulations, providing concrete scenarios for better understanding and decision making. Compared to numerical models for ensemble generation, our method could provide a much less computationally expensive alternative.

\clearpage

\bibliography{example_paper}
\bibliographystyle{icml2024}

\newpage
\appendix
\onecolumn
\section{Variational Autoencoder}
\label{appendix1}

The Variational Autoencoder (VAE) is based on Rombach et al. \citeyearpar{rombach2022high} and focuses on perceptual image compression. The encoder (\(E\)) encodes an image \( x \in \mathbb{R}^{H \times W \times 3} \) into a latent representation \( z = E(x) \), where \( z \in \mathbb{R}^{h \times w \times c} \). The decoder \(D\) reconstructs the image from the latent representation, giving \( \tilde{x} = D(z) = D(E(x)) \). The compression factor is given by \( f = \frac{H}{h} = \frac{W}{w} \). In our setup, we use a compression factor of $f = 8$ and a latent dimension of $c = 4$, giving us a total compression of 16.

The given objective $\mathcal{L}_{\text{VAE}}$ combines the reconstruction loss, adversarial loss, and Kullback-Leibler divergence (KL-divergence) regularization for training the VAE with adversarial training. The reconstruction loss $\mathcal{L}_{\text{rec}}$ measures the L2 distance between the original image $x$ and the reconstructed image $\tilde{x}$ to produce images that are close to the input in pixel space:

\begin{equation}
\mathcal{L}_{\text{rec}} = \| x - \tilde{x} \|_2^2
\end{equation}

The adversarial loss ensures that the reconstructed images are perceptually similar to the real images by adding a discriminator $D$:

\begin{equation}
\mathcal{L}_{\text{adv}} = \mathbb{E}_{x \sim p_{\text{data}}(x)} [\log D(x)] + \mathbb{E}_{\tilde{x} \sim p_{\text{model}}(\tilde{x})} [\log (1 - D(\tilde{x}))]
\end{equation}

The KL divergence regularizes the latent space by making the distribution of the encoded latent variables $q(z|x)$ close to a prior distribution $p(z)$, in our case a standard normal distribution. The overall objective for the VAE with adversarial training is given by:

\begin{equation}
\mathcal{L}_{\text{VAE}} = \mathbb{E}_{x \sim p_{\text{data}}(x)} \left[ \lambda_{\text{rec}} \| x - \tilde{x} \|_2^2 + \lambda_{\text{adv}} \left( \mathbb{E}_{x \sim p_{\text{data}}(x)} [\log D(x)] + \mathbb{E}_{\tilde{x} \sim p_{\text{model}}(\tilde{x})} [\log (1 - D(\tilde{x}))] \right) + \lambda_{\text{KL}} \text{KL}(q(z|x) \| p(z)) \right]
\end{equation}

\begin{figure}[ht]
\begin{center}
\centerline{\includegraphics[width=.55\columnwidth]{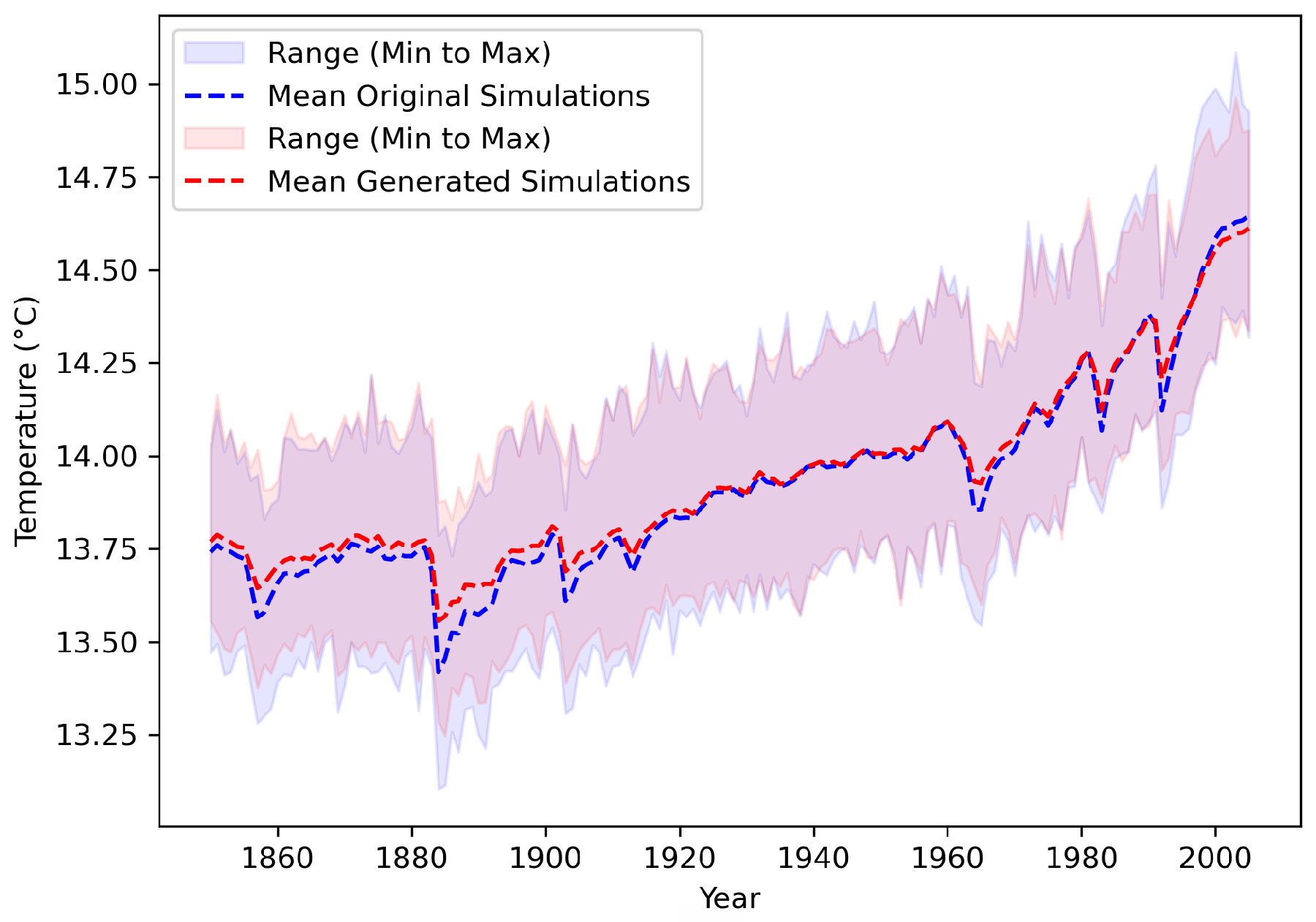}}
\caption{Ensemble spread and ensemble mean of annual spatially averaged 98 original members from the MPI-GE (blue) compared to the reconstructed members of the VAE (red) ranging from 1850 to 2005.}
\label{vae-mean-spread}
\end{center}
\end{figure}

We evaluated the performance of our VAE independently of the overall setup to ensure that the diffusion model was provided with high quality latent data. To do this, we mapped our complete data set of simulations into latent space using the encoder $E$ and reconstructed all the data using the decoder $D$. Our VAE achieved a pixel-wise RMSE of $0.25$°C. Figure \ref{vae-mean-spread} also shows the comparison of the spatial annual mean and spread from the original and reconstructed datasets over the full time range. 

\newpage

\section{Denoising Diffusion Model}
\label{appendix2}
Diffusion models generate images by reversing a gradual noise process \cite{sohl2015deep}. The process starts with pure noise and iteratively denoises it to produce a high-quality image. The main steps in this process are a forward and a backward diffusion process. The forward process gradually adds Gaussian noise to the data over a series of diffusion steps. Starting from the original image \(x_0\), noise is added at each diffusion step \(t\) to produce a noisy image \(x_t\). This process can be described by the following equation, where \(\beta_t\) is a variance schedule that controls the amount of noise added at each step:
\begin{equation}
     q(x_t | x_{t-1}) = \mathcal{N}(x_t; \sqrt{1-\beta_t} x_{t-1}, \beta_t I)
\end{equation}
The reverse diffusion process is learned by training a neural network to predict the noise component added at each step. The model \( \epsilon_\theta \) predicts the noise given the noisy image \(x_t\) and the diffusion step \(t\). The objective is to minimise the difference between the predicted noise and the true noise, in our case using a mean square error loss:
\begin{equation}
     \mathcal{L} = \mathbb{E}_{t, x_0, \epsilon} \left[ \| \epsilon - \epsilon_\theta(x_t, t) \|^2 \right]
\end{equation}
The reverse process can then generate \(x_{t-1}\) from \(x_t\) using the learned model, where \(\mu_\theta\) and \(\Sigma_\theta\) are functions of the predicted noise:
\begin{equation}
    p_\theta(x_{t-1} | x_t) = \mathcal{N}(x_{t-1}; \mu_\theta(x_t, t), \Sigma_\theta(x_t, t))
\end{equation}

Our model is based on the work of Dhariwal \& Nichol \citeyearpar{dhariwal2021diffusion} and implements a UNet architecture with residual blocks as described by Brock et al. \cite{brock2018large}. Similar to the original implementation used for image synthesis, we apply spatial attention mechanisms in the upper layers of the model. To address the time cost associated with iterative inference over large data sequences, our model operates in the latent space. For autoregressive prediction sequence processing, the temporal dimension of the latent simulations is encoded within the channel dimension of the model.

The transformer-based component of our model processes temporal information in an additional dimension. Each transformer block consists of three parts: a spatial attention mechanism, a temporal attention mechanism, and a multilayer perceptron (MLP). The core concept of the transformer \cite{vaswani2017attention} is the self-attention mechanism, which allows the model to evaluate the importance of different regions in the spatial dimension or time steps in the temporal dimension. The data is divided into patches (either spatial or temporal) and transformed into three vectors: Query (Q), Key (K) and Value (V). The size of the temporal patches varies with the depth of the network; higher layers consider smaller timescales, while the bottleneck layer includes all timescales. Attention scores are computed by taking the dot product of the query vector with all key vectors, followed by a softmax function to obtain weights. These weights are then used to compute a weighted sum of the value vectors:

\begin{equation}
\text{Attention}(Q, K, V) = \text{softmax}\left(\frac{QK^T}{\sqrt{d_k}}\right)V
\end{equation}

Instead of performing a single attention function, the transformer uses multiple attention heads to capture different aspects of the relationships between elements. Each head has its own set of $Q$, $K$ and $V$ matrices, and their outputs are concatenated and linearly transformed. In our model, the number of attention heads increases with layer depth, starting with fewer heads in the early layers and reaching a maximum number in the bottleneck layer. Following the multi-head attention mechanisms, a residual connection MLP network is applied. This consists of a layer normalisation, a linear layer and a Gaussian Error Linear Unit (GELU) activation function.

\newpage

\section{Extended Analysis}
\label{appendix3}
\begin{figure}[ht]
\begin{center}
\centerline{\includegraphics[width=.55\columnwidth]{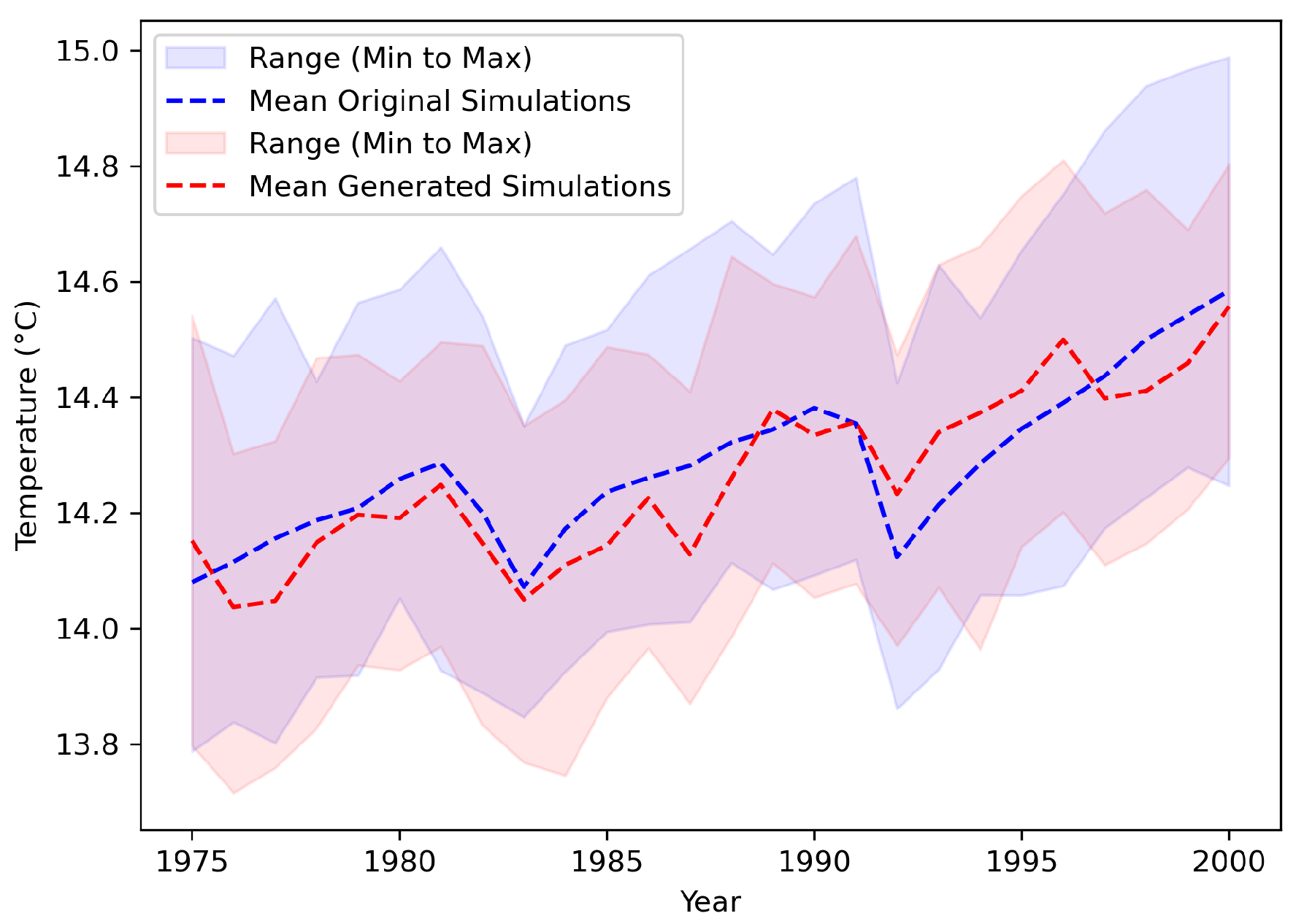}}
\caption{Ensemble spread and ensemble mean of annual spatially averaged 100 original members from the MPI-GE (blue) compared to the reconstructed members of the autoregressive model (red) ranging from 1975 to 2000.}
\label{future-mean-spread}
\end{center}
\end{figure}

We investigated the generalisability of our model to unseen time periods. The channel-based autoregressive diffusion model was trained on all historical MPI-GE members from 1850 to 1975. We then conditioned the trained model on a single simulation from 1975 to 2000, generating 100 members. Figure \ref{future-mean-spread} shows the results. While the mean and spread of the generated simulations do not perfectly match the original ones, the simulations successfully capture the ongoing global warming trend despite not being trained on this period. In addition, the generated simulations strongly reflect the major volcanic eruptions of 1982 and 1991.

\begin{figure}[ht]
\begin{center}
\centerline{\includegraphics[width=\columnwidth]{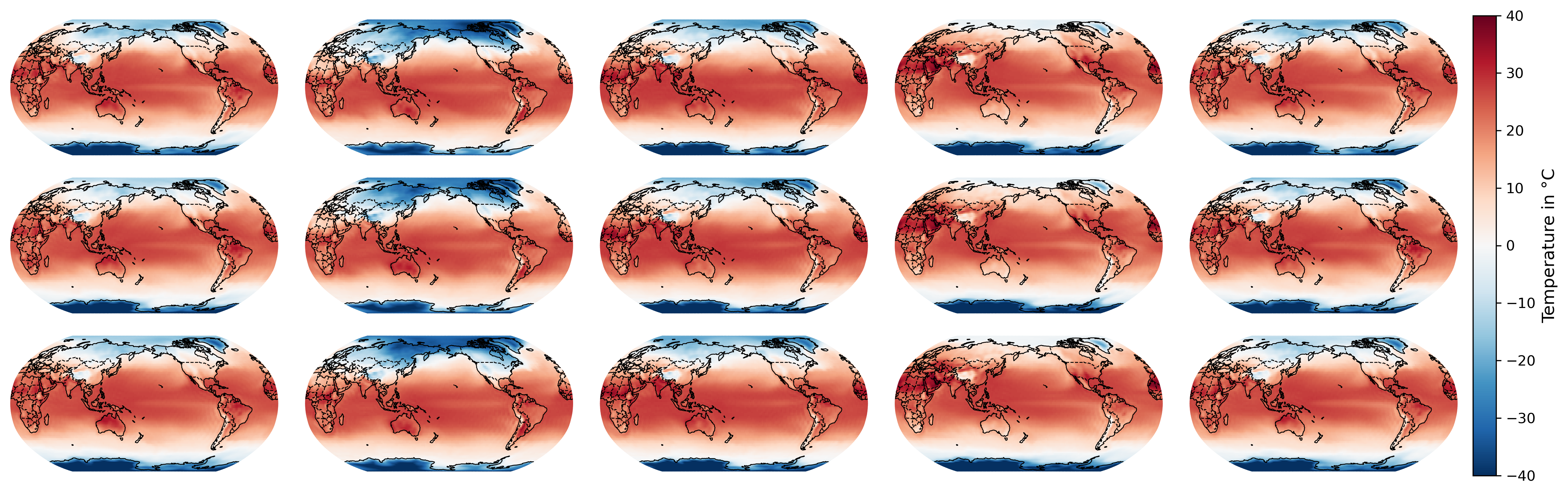}}
\caption{Absolute temperature maps of some randomly chosen samples. Top row shows the samples from an original MPI-GE simulation, center row from the autoregressive technique and bottom row from the transformer-based technique.}
\label{temperature-maps}
\end{center}
\end{figure}

\end{document}

%% file: ddpm.pdf_tex
\begingroup%
  \makeatletter%
  \providecommand\color[2][]{%
    \errmessage{(Inkscape) Color is used for the text in Inkscape, but the package 'color.sty' is not loaded}%
    \renewcommand\color[2][]{}%
  }%
  \providecommand\transparent[1]{%
    \errmessage{(Inkscape) Transparency is used (non-zero) for the text in Inkscape, but the package 'transparent.sty' is not loaded}%
    \renewcommand\transparent[1]{}%
  }%
  \providecommand\rotatebox[2]{#2}%
  \newcommand*\fsize{\dimexpr\f@size pt\relax}%
  \newcommand*\lineheight[1]{\fontsize{\fsize}{#1\fsize}\selectfont}%
  \ifx\svgwidth\undefined%
    \setlength{\unitlength}{232.26717653bp}%
    \ifx\svgscale\undefined%
      \relax%
    \else%
      \setlength{\unitlength}{\unitlength * \real{\svgscale}}%
    \fi%
  \else%
    \setlength{\unitlength}{\svgwidth}%
  \fi%
  \global\let\svgwidth\undefined%
  \global\let\svgscale\undefined%
  \makeatother%
  \begin{picture}(1,0.5840734)%
    \lineheight{1}%
    \setlength\tabcolsep{0pt}%
    \put(0,0){\includegraphics[width=\unitlength,page=1]{ddpm.pdf}}%
    \put(0.38358795,0.37339195){\color[rgb]{0,0,0}\makebox(0,0)[t]{\lineheight{1.25}\smash{\begin{tabular}[t]{c}\textsl{E}\end{tabular}}}}%
    \put(0,0){\includegraphics[width=\unitlength,page=2]{ddpm.pdf}}%
    \put(0.5688791,0.37339195){\color[rgb]{0,0,0}\makebox(0,0)[t]{\lineheight{1.25}\smash{\begin{tabular}[t]{c}\textsl{D}\end{tabular}}}}%
    \put(0.4800303,0.46207515){\color[rgb]{0,0,0}\makebox(0,0)[t]{\lineheight{1.25}\smash{\begin{tabular}[t]{c}VAE\end{tabular}}}}%
    \put(0,0){\includegraphics[width=\unitlength,page=3]{ddpm.pdf}}%
    \put(0.47928405,0.10186823){\color[rgb]{0,0,0}\makebox(0,0)[t]{\lineheight{1.25}\smash{\begin{tabular}[t]{c}DDM\end{tabular}}}}%
    \put(0,0){\includegraphics[width=\unitlength,page=4]{ddpm.pdf}}%
    \put(0.81803721,0.38140752){\color[rgb]{0,0,0}\makebox(0,0)[t]{\lineheight{1.25}\smash{\begin{tabular}[t]{c}$\hat{x}=[\hat{x}_0,...,\hat{x}_t]$\end{tabular}}}}%
    \put(0.13099554,0.38140752){\color[rgb]{0,0,0}\makebox(0,0)[t]{\lineheight{1.25}\smash{\begin{tabular}[t]{c}${x}_c=[{x}_0,...,{x}_t]$\end{tabular}}}}%
    \put(0,0){\includegraphics[width=\unitlength,page=5]{ddpm.pdf}}%
    \put(0.48118907,0.55264499){\color[rgb]{0,0,0}\makebox(0,0)[t]{\lineheight{1.25}\smash{\begin{tabular}[t]{c}\scriptsize{training on single climate states}\end{tabular}}}}%
    \put(0.38745849,0.2286129){\color[rgb]{0,0,0}\makebox(0,0)[t]{\lineheight{1.25}\smash{\begin{tabular}[t]{c}${z}_c$\end{tabular}}}}%
    \put(0,0){\includegraphics[width=\unitlength,page=6]{ddpm.pdf}}%
    \put(0.57151372,0.22997372){\color[rgb]{0,0,0}\makebox(0,0)[t]{\lineheight{1.25}\smash{\begin{tabular}[t]{c}$\hat{z}_y$\end{tabular}}}}%
    \put(0,0){\includegraphics[width=\unitlength,page=7]{ddpm.pdf}}%
    \put(0.47513527,0.00952312){\color[rgb]{0,0,0}\makebox(0,0)[t]{\lineheight{1.25}\smash{\begin{tabular}[t]{c}\scriptsize{training on sequences}\end{tabular}}}}%
  \end{picture}%
\endgroup%

%% file: example_paper.bbl
\begin{thebibliography}{16}
\providecommand{\natexlab}[1]{#1}
\providecommand{\url}[1]{\texttt{#1}}
\expandafter\ifx\csname urlstyle\endcsname\relax
  \providecommand{\doi}[1]{doi: #1}\else
  \providecommand{\doi}{doi: \begingroup \urlstyle{rm}\Url}\fi

\bibitem[Brochet et~al.(2023)Brochet, Raynaud, Thome, Plu, and Rambour]{brochet2023multivariate}
Brochet, C., Raynaud, L., Thome, N., Plu, M., and Rambour, C.
\newblock Multivariate emulation of kilometer-scale numerical weather predictions with generative adversarial networks: A proof of concept.
\newblock \emph{Artificial Intelligence for the Earth Systems}, 2\penalty0 (4):\penalty0 230006, 2023.

\bibitem[Brock et~al.(2018)Brock, Donahue, and Simonyan]{brock2018large}
Brock, A., Donahue, J., and Simonyan, K.
\newblock Large scale gan training for high fidelity natural image synthesis.
\newblock \emph{arXiv preprint arXiv:1809.11096}, 2018.

\bibitem[Dhariwal \& Nichol(2021)Dhariwal and Nichol]{dhariwal2021diffusion}
Dhariwal, P. and Nichol, A.
\newblock Diffusion models beat gans on image synthesis.
\newblock \emph{Advances in neural information processing systems}, 34:\penalty0 8780--8794, 2021.

\bibitem[Ho et~al.(2020)Ho, Jain, and Abbeel]{ho2020denoising}
Ho, J., Jain, A., and Abbeel, P.
\newblock Denoising diffusion probabilistic models.
\newblock \emph{Advances in neural information processing systems}, 33:\penalty0 6840--6851, 2020.

\bibitem[Kingma \& Welling(2013)Kingma and Welling]{kingma2013auto}
Kingma, D.~P. and Welling, M.
\newblock Auto-encoding variational bayes.
\newblock \emph{arXiv preprint arXiv:1312.6114}, 2013.

\bibitem[Lorenz et~al.(2018)Lorenz, Herger, Sedl{\'a}{\v{c}}ek, Eyring, Fischer, and Knutti]{lorenz2018prospects}
Lorenz, R., Herger, N., Sedl{\'a}{\v{c}}ek, J., Eyring, V., Fischer, E.~M., and Knutti, R.
\newblock Prospects and caveats of weighting climate models for summer maximum temperature projections over north america.
\newblock \emph{Journal of Geophysical Research: Atmospheres}, 123\penalty0 (9):\penalty0 4509--4526, 2018.

\bibitem[Maher et~al.(2019)Maher, Milinski, Suarez-Gutierrez, Botzet, Dobrynin, Kornblueh, Kr{\"o}ger, Takano, Ghosh, Hedemann, et~al.]{maher2019max}
Maher, N., Milinski, S., Suarez-Gutierrez, L., Botzet, M., Dobrynin, M., Kornblueh, L., Kr{\"o}ger, J., Takano, Y., Ghosh, R., Hedemann, C., et~al.
\newblock The max planck institute grand ensemble: enabling the exploration of climate system variability.
\newblock \emph{Journal of Advances in Modeling Earth Systems}, 11\penalty0 (7):\penalty0 2050--2069, 2019.
\newblock \doi{10.1029/2019MS001639}.

\bibitem[Price et~al.(2023)Price, Sanchez-Gonzalez, Alet, Ewalds, El-Kadi, Stott, Mohamed, Battaglia, Lam, and Willson]{price2023gencast}
Price, I., Sanchez-Gonzalez, A., Alet, F., Ewalds, T., El-Kadi, A., Stott, J., Mohamed, S., Battaglia, P., Lam, R., and Willson, M.
\newblock Gencast: Diffusion-based ensemble forecasting for medium-range weather.
\newblock \emph{arXiv preprint arXiv:2312.15796}, 2023.

\bibitem[Rasul et~al.(2021)Rasul, Seward, Schuster, and Vollgraf]{rasul2021autoregressive}
Rasul, K., Seward, C., Schuster, I., and Vollgraf, R.
\newblock Autoregressive denoising diffusion models for multivariate probabilistic time series forecasting.
\newblock In \emph{International Conference on Machine Learning}, pp.\  8857--8868. PMLR, 2021.

\bibitem[Reichstein et~al.(2019)Reichstein, Camps-Valls, Stevens, Jung, Denzler, Carvalhais, and Prabhat]{reichstein2019deep}
Reichstein, M., Camps-Valls, G., Stevens, B., Jung, M., Denzler, J., Carvalhais, N., and Prabhat.
\newblock Deep learning and process understanding for data-driven earth system science.
\newblock \emph{Nature}, 566\penalty0 (7743):\penalty0 195--204, 2019.

\bibitem[Rombach et~al.(2022)Rombach, Blattmann, Lorenz, Esser, and Ommer]{rombach2022high}
Rombach, R., Blattmann, A., Lorenz, D., Esser, P., and Ommer, B.
\newblock High-resolution image synthesis with latent diffusion models.
\newblock In \emph{Proceedings of the IEEE/CVF conference on computer vision and pattern recognition}, pp.\  10684--10695, 2022.

\bibitem[Ronneberger et~al.(2015)Ronneberger, Fischer, and Brox]{ronneberger2015u}
Ronneberger, O., Fischer, P., and Brox, T.
\newblock U-net: Convolutional networks for biomedical image segmentation.
\newblock In \emph{Medical image computing and computer-assisted intervention--MICCAI 2015: 18th international conference, Munich, Germany, October 5-9, 2015, proceedings, part III 18}, pp.\  234--241. Springer, 2015.

\bibitem[Sohl-Dickstein et~al.(2015)Sohl-Dickstein, Weiss, Maheswaranathan, and Ganguli]{sohl2015deep}
Sohl-Dickstein, J., Weiss, E., Maheswaranathan, N., and Ganguli, S.
\newblock Deep unsupervised learning using nonequilibrium thermodynamics.
\newblock In \emph{International conference on machine learning}, pp.\  2256--2265. PMLR, 2015.

\bibitem[Song et~al.(2020)Song, Meng, and Ermon]{song2020denoising}
Song, J., Meng, C., and Ermon, S.
\newblock Denoising diffusion implicit models.
\newblock \emph{arXiv preprint arXiv:2010.02502}, 2020.

\bibitem[Trenberth(1997)]{trenberth1997definition}
Trenberth, K.~E.
\newblock The definition of el nino.
\newblock \emph{Bulletin of the American Meteorological Society}, 78\penalty0 (12):\penalty0 2771--2778, 1997.

\bibitem[Vaswani et~al.(2017)Vaswani, Shazeer, Parmar, Uszkoreit, Jones, Gomez, Kaiser, and Polosukhin]{vaswani2017attention}
Vaswani, A., Shazeer, N., Parmar, N., Uszkoreit, J., Jones, L., Gomez, A.~N., Kaiser, {\L}., and Polosukhin, I.
\newblock Attention is all you need.
\newblock \emph{Advances in neural information processing systems}, 30, 2017.

\end{thebibliography}
